\documentclass{article}

\usepackage{arxiv}

\usepackage[utf8]{inputenc} % allow utf-8 input
\usepackage[T1]{fontenc}    % use 8-bit T1 fonts
\usepackage{hyperref}       % hyperlinks
\usepackage{url}            % simple URL typesetting
\usepackage{booktabs}       % professional-quality tables
\usepackage{amsfonts}       % blackboard math symbols
\usepackage{nicefrac}       % compact symbols for 1/2, etc.
\usepackage{microtype}      % microtypography
\usepackage{lipsum}		% Can be removed after putting your text content
\usepackage{graphicx}
\usepackage{natbib}
\usepackage{doi}

\title{GENERATIVE PRE-TRAINED TRANSFORMER FOR DESIGN CONCEPT GENERATION: AN EXPLORATION }

\date{November 15, 2021}	% Here you can change the date presented in the paper title
\author{ \href{https://orcid.org/0000-0002-5401-6679}{\includegraphics[scale=0.06]{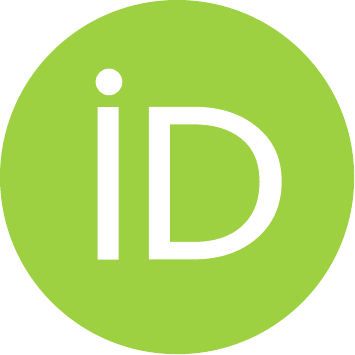}\hspace{1mm}Qihao Zhu}\\
	Data-Driven Innovation Lab\\
	Singapore University of Technology and Design\\
	\texttt{qihao\_zhu@mymail.sutd.edu.sg} \\
	\And
	\href{https://orcid.org/0000-0001-5892-8432}{\includegraphics[scale=0.06]{orcid.pdf}\hspace{1mm}Jianxi Luo} \\
	Data-Driven Innovation Lab\\
	Singapore University of Technology and Design\\
	\texttt{jianxi\_luo@sutd.edu.sg} \\
}

\renewcommand{\shorttitle}{Generative Pre-Trained Transformer for Design Concept Generation: An Exploration}

\begin{document}
\maketitle

\begin{abstract}
Novel concepts are essential for design innovation and can be generated with the aid of data stimuli and computers. However, current generative design algorithms focus on diagrammatic or spatial concepts that are either too abstract to understand or too detailed for early phase design exploration. This paper explores the uses of generative pre-trained transformers (GPT) for natural language design concept generation. Our experiments involve the use of GPT-2 and GPT-3 for different creative reasonings in design tasks. Both show reasonably good performance for verbal design concept generation.
\end{abstract}

% keywords can be removed

\section{Introduction}
Design innovation is heavily dependent on high-quality and novel concepts. Concept design activities are divided into two stages: divergence and convergence \citep{tschimmel2012design}. Designers must generate a wide range of concepts during the divergence stage before any assessment and selection for convergence to be made. Therefore, much research has been conducted to develop methodologies and tools to aid designers to create design concepts, often with the aid of data and computers \citep{chakrabarti2011computer,han2018combinator,han2020data}.\par
Meanwhile, little progress has been made regarding "computer ideation", i.e., computers directly and automatically generating ideas, in contrast to computer-aided ideation, i.e., computers aiding or stimulating human designers to generate ideas. In this study, we introduce a new technique from the field of artificial intelligence (AI)--the generative pre-trained transformer (GPT)--for automated generation of verbal design concepts. GPTs are language models pre-trained on vast quantities of textual data and can perform a wide range of language-related tasks \citep{radford2019language,brown2020language}. Our work experiments the applicability of different GPT models for both problem-driven reasoning and analogy-driven reasoning for design, with customized datasets.

\section{Literature Review}
\label{sec:headings}
Recent research on design concept generation can be categorized based on three dimensions: the role of method or tool, the form of concept representation, and the targeted design process stage.\par
A concept generation method or tool can play one of the three roles: as a guide, as a stimulator, or as a generator. A method is considered as guide if it is instructively involved in the generation of design concepts, providing design rules or guidelines to the activities of the designers. For instance, \citet{bonnardel2020brainstorming} proposed two variants of brainstorming, encouraging designers to focus on the evocation of both the design ideas and the constraints related to the design problem. A stimulator provides inspirational stimuli to provoke designers to conceive new concepts. \citet{goldschmidt2006variances} discussed how visual stimuli affect problem solving design performance. \citet{jin2020new} extracted 10 design heuristics as stimuli from RedDot award-wining design concepts to help digital designers overcome design fixation. \citet{he2019mining} tested the use of word clouds as stimulators to inspire ideation. Meanwhile, some methods can be a guide and a simulator at the same time. \citet{luo2019computer} introduced a computer-aided ideation tool InnoGPS to guide the provision of design stimuli from the patent database by their knowledge distance to the design problem or interest. \citet{fargnoli2006morphological} introduced the morphological matrix to guide designers to navigate and combine alternative solutions to each of multiple functions of a product to generate a variety of designs.\par
A concept generator represents a fully automated computational agent that creates new concepts for the interest of designers. One example is function-based design synthesis \citep{chakrabarti2011computer}. \citet{sangelkar2017automated} introduced graph grammar to generate the function structures of the design concepts with potential for computational design synthesis. \citet{kang2015automated} proposed a concept generation method based on function-form synthesis. Other generators perform topology optimization and generative visual design \citep{vlah2020evaluation}. \citet{oh2019deep} and \citet{nie2021topologygan} integrate topology optimization and generative adversarial network (GAN) to generate concepts for both aesthetic and engineering performance. \citet{ren2013quantification}, \citet{burnap2016estimating}, and \citet{dogan2019generative} use generative models to create new concepts of vehicle form design.\par
During design activities, concepts can be represented in the forms of abstract diagram, verbal text, or spatial visualization. Guiding or stimulation-based methods can direct designers to generate concepts in either of the three forms, e.g., simple sketches \citep{shah2001collaborative,goldschmidt2006variances}, mind-mapping graph \citep{shih2009groupmind,yagita2011validation}, functional diagram \citep{stone2000heuristic}, or textual description \citep{he2019mining,sarica2021idea}. Some guides or stimulators may also lead to multiple forms of concept representation as designers may record their perception of ideas in different ways \citep{bonnardel2020brainstorming,ilevbare2013review,yilmaz2016evidence}. On the other hand, for automated tools, the type of concepts to be generated is pre-determined when designing the system, e.g., the graph grammar-based tools represent new concepts in abstract graphs \citep{campbell2009graph,sangelkar2017automated}, while the topology optimization tools generate spatially visualized concepts in the form of 2D images \citep{oh2019deep} or 3D models \citep{nie2021topologygan}.\par
The forms of generated concepts need to fit with design process stages. \citet{pahl2007engineering} specified four stages in design processes, including planning and task clarification, conceptual design, embodiment design, and detail design. For designers, visual stimulators like mood boards \citep{ahmed2006investigation} and spatial concept generators \citep{oh2019deep,nie2021topologygan} are mainly useful for embodiment and detail design stages and may cause design fixation if applied in earlier stages \citep{viswanathan2016study}. Diagrammatic concepts are represented in a more abstract way that either visualizes the mind map of a design concept \citep{shih2009groupmind,yagita2011validation} or the relationship between components of function or structure \citep{stone2000heuristic,campbell2009graph}. They are more suitable for the planning and task clarification as well as conceptual design stages. Meanwhile, text is also a common modality of recording concepts for early design stages. In typical brainstorming sessions, designers exchange preliminary design ideas verbally and frequently. \citet{chiu2007understanding} investigated how semantic stimuli presented as words affect concept generation. \citet{sarica2021idea} retrieved the terms from a pre-trained technology semantic network as stimuli to generate new concepts in the form of text. \citet{goucher2019crowdsourcing} and \citet{camburn2020machine} collected design ideas written in short text from crowdsourcing campaigns. To date, however, there exist no automated tool that is built to generate verbal design concepts.\par
To summarize, we present a taxonomy of concept generation methods or tools in Figure 1 by their roles, concept representation forms and suitable design stages. In this research, we focus on verbal generator using the latest natural language generation (NLG) technology. Particularly, we experiment the generative pre-trained transformers from OpenAI to learn design knowledge and reasoning from task-oriented datasets and then generate high fidelity design concept descriptions in natural language.
%\lipsum[4] See Section \ref{sec:headings}.
%\graphicspath{{./figure1 ver2.jpg/}}
\begin{figure}
	\centering
	\includegraphics[width = 12cm]{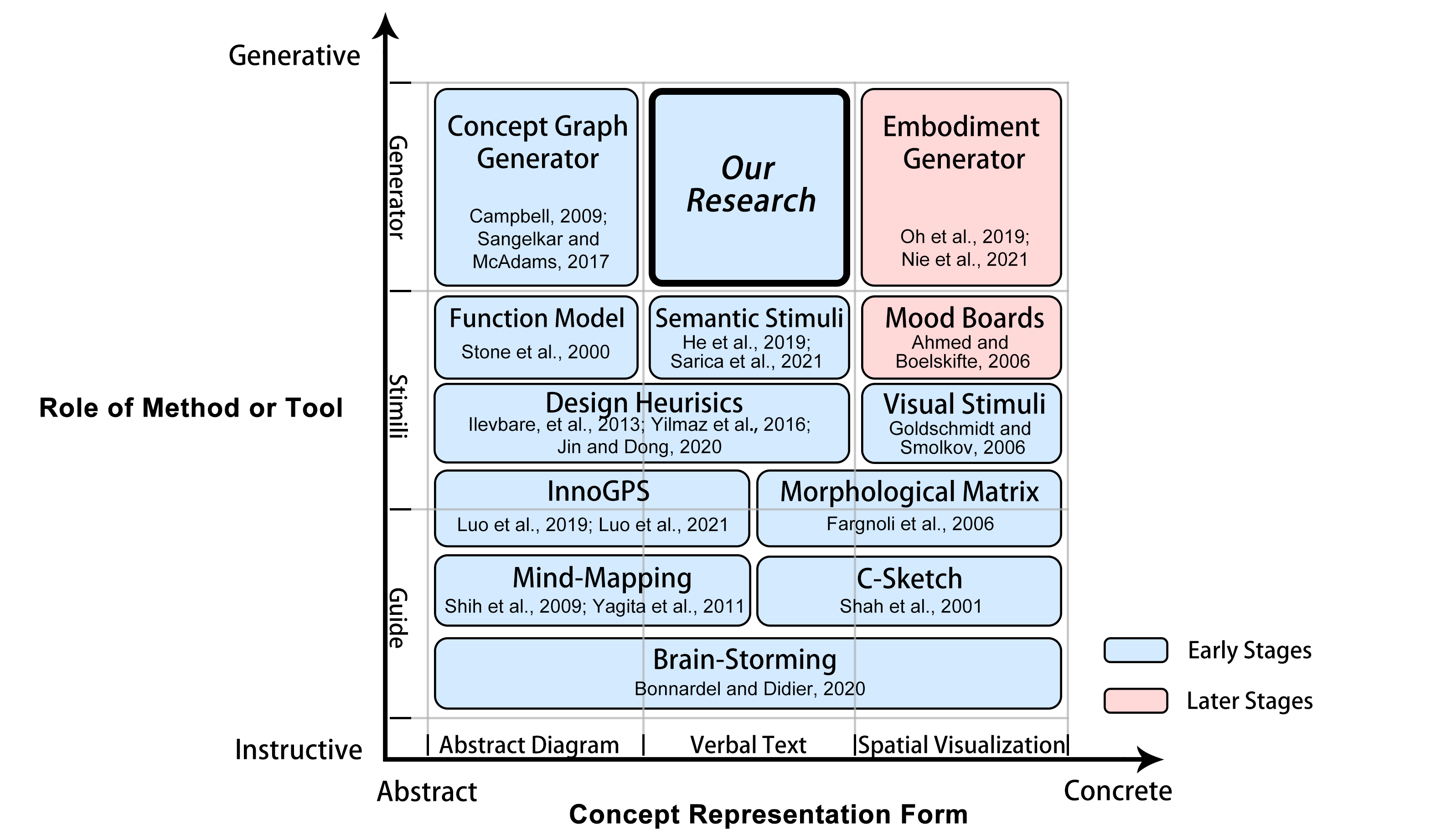}
	%\fbox{\rule[-.5cm]{4cm}{4cm} \rule[-.5cm]{4cm}{0cm}}
	\caption{Taxonomy of concept generation methods or tools}
	\label{fig:fig1}
\end{figure}

\section{Natural Language Generation (NLG)}
\label{sec:headings}
\subsection{Text-to-Text and Data-to-Text NLG}
Natural language generation (NLG) is considered as a computer program that generates natural language as output \citep{gatt2018survey}. According to \citet{gatt2018survey}, depending on the input, there are two major instances of NLG: text-to-text and data-to-text generation. Text-to-text is what takes in existing text as input and then outputs new pieces of text. Examples of text-to-text generation include machine translation (e.g., \citet{kenny2019machine}), text summarization or simplification (e.g., \citet{ozsoy2011text}), automatic writing correction (e.g., \cite{karyuatry2018grammarly}), paraphrasing (e.g., \citet{li2018paraphrase}), and so on. On the other hand, data-to-text generation takes in non-linguistic data as input. Applications of data-to-text have been seen in varied fields, e.g., generating news based on election data \citep{leppanen2017data}, generating personalized suggestions based on user input and web data \citep{wanner2015getting}. 
%\begin{equation}
%	\xi _{ij}(t)=P(x_{t}=i,x_{t+1}=j|y,v,w;\theta)= {\frac {\alpha _{i}(t)a^{w_t}_{ij}\beta _{j}(t+1)b^{v_{t+1}}_{j}(y_{t+1})}{\sum _{i=1}^{N} \sum _{j=1}^{N} \alpha _{i}(t)a^{w_t}_{ij}\beta _{j}(t+1)b^{v_{t+1}}_{j}(y_{t+1})}}
%\end{equation}
\subsection{Transformer for NLG}
Transformer, first introduced by \citet{vaswani2017attention}), is the state-of-the-art neural network architecture for natural language processing (NLP) and is becoming the dominant for NLG \citep{topal2021exploring}. Comparing to recurrent neural network (RNN) and long short-term memory (LSTM), which were the most popular neural network architectures until recently, transformer overcomes the vanishing gradient problem \citep{pascanu2013difficulty}, which could cause the failure of maintaining context when processing longer text. Moreover, transformer enables parallel training. With the training data and model architecture become larger in size, it can capture longer sequence features and therefore result in much more comprehensive language understanding and generation \citep{brown2020language}.\par
Another significant feature of transformer for NLG is that it blurs the boundary between data-to-text and text-to-text NLG models. For instance, \citet{radford2019language} reports the outstanding performance of the GPT-2 transformer model from OpenAI for text-to-text generation tasks such as translation and summarization, meanwhile \citet{peng2020few} shows the model after fine-tuning is also capable of performing data-to-text dialogue generation tasks.\par
Although transformer is a rather new technique in NLP and NLG, some applications have already been seen in different fields. \citet{amin2020exploring} use transformer models to generate structured patient information to augment medical dataset. \citet{lee2020patent} use GPT-2 with fine-tuning to generate patent claims. \citet{fang2021application} also use GPT to generate ideas for content creators. However, according to a recent review conducted by \citet{regenwetter2021deep}, the application of transformers is still a wide-open space for engineering design tasks. Our work fills this gap.

\subsection{Generative Pre-trained Transformer (GPT)}
Now we introduce the most popular series of transformers in NLG tasks: the generative pre-trained transformer \citep{radford2019language,brown2020language}, or GPT for short. Later in this paper, we apply and compare the 2nd and 3rd generations of GPT, i.e., GPT-2 and GPT-3, in design concept generation tasks.
GPT-2 uses the two-step training strategy of pre-training and fine-tuning, following \citet{hinton2006reducing}. The workflow is shown in Figure 2(a). During the pre-training step, the model is trained on a massive text dataset called WebText, which is collected from millions of webpages \citep{radford2019language}. This results in a comprehensive language model that can perform general language completion tasks. According to OpenAI, the largest GPT-2 pre-trained model has 1.5 billion parameters. For downstream NLP tasks, the pre-trained model then needs to be fine-tuned given a customized and task-oriented dataset. The fine-tuned model is trained through repeated gradient updates using a large dataset of corpus of the example task. This process updates the weights of the pre-trained model and stores them for the use of the target task. However, the large dataset suitable for the target NLP task may be unavailable or difficult to collect.\par
GPT-3 is the largest language model so far. It is trained on a mixture of datasets containing 400 billion tokens and has a maximum of 175 billion parameters, over a hundred times larger than GPT-2. Comparing to its precursor, GPT-3 can perform a wide range of NLP tasks without the need of fine-tuning. GPT-3 is capable of few-show learning \citep{brown2020language}, in which the model learns from multiple examples of the NLP or NLG task before conducting it. In this process, no gradient updates are performed \citep{brown2020language}. The training process of GPT-3 is shown in Figure 2 (b), where the prompts are the task examples given to the model for few-shot learning.\par
For text generation, the implementation of GPT requires several control parameters and can be categorized in three types according to the aimed property of the generated text. The ‘max\_tokens’ (also called ‘length’), ‘prompt’ (also called ‘profix’), and ‘stop’ (also called ‘truncate’) control the content and format of the text. These are essential parameters when customizing the task and defining the input of the NLG model. When conditionally sampling with a pre-determined prompt, the model will learn to set up the context and output results based on it. The ‘temperature’, ‘top-k’ (only applicable for GPT-2), and ‘top-p’ parameters control the randomness of the generated text. Higher randomness will result in more varied outputs. Finally, the ‘presence\_penalty’ and ‘frequency\_penalty’ parameters are only applicable for GPT-3 and control generation repetitiveness. By encouraging new topics and discouraging existing ones, the generated texts are more likely to deviate from the given examples and represent novel concepts.

\begin{figure}
	\centering
	\includegraphics[width = 16cm]{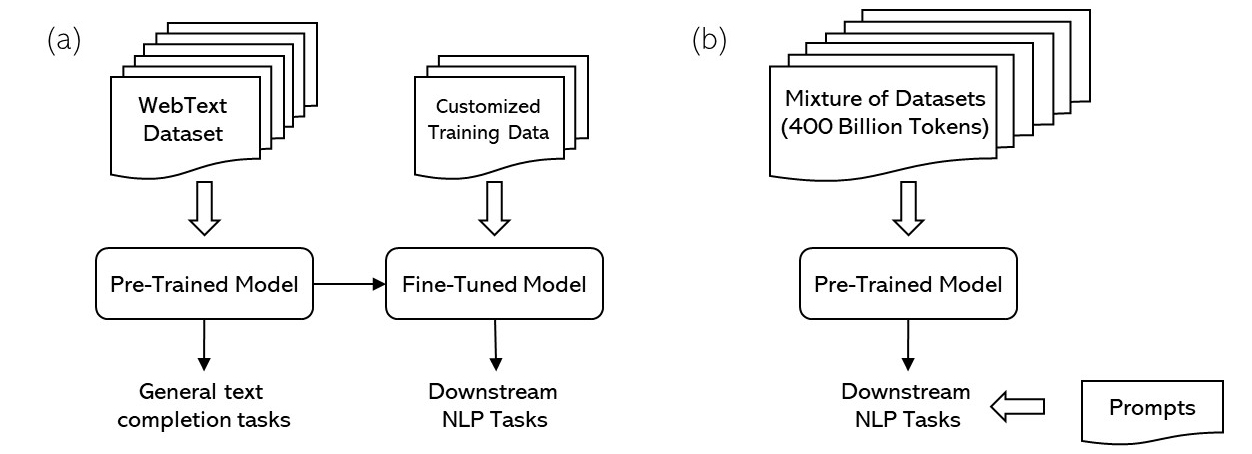}
	%\fbox{\rule[-.5cm]{4cm}{4cm} \rule[-.5cm]{4cm}{0cm}}
	\caption{Training and re-training process of GPT-2 (a) and GPT-3 (b)}
	\label{fig:fig2}
\end{figure}

%\subsubsection{Headings: third level}
%\lipsum[6]

\section{Research Method}
\label{sec:headings}
In this paper, we experiment the applications of different GPT models in different design concept generation tasks. Figure 3 depicts the general framework of our experiments. First, knowledge for the task is the key component of our framework. It is provided through the dataset used for fine-tuning GPT-2 model, or in the examples of design concepts for few-shot learning of GPT-3. Secondly, we take in varied input as the prompt for conditional learning. The input should be customized and consistent with the specific reasoning we want the model to learn, and output will be the generated design concept description. For instance, for analogy reasoning, the input will be the source and target domains for analogy mapping. Finally, the transformer in the framework can be either a fine-tuned GPT-2 or the pre-trained GPT-3 for few-shot learning.\par
Table 1 summarizes the settings for two experiments, in which we explore the capability of GPT for generating concepts by problem-driven reasoning and analogy-driven reasoning. The data we use for knowledge acquirement in both experiments is from the repository of RedDot award-wining designs. The first experiment includes three phrases for implementation: preparing the data for the NLG task, fine-tuning the model with the provided dataset, and testing the performance. The fine-tuning phrase is not included for the second experiment because we will be employing GPT-3 few-shot learning.

\begin{figure}
	\centering
	\includegraphics[width = 8cm]{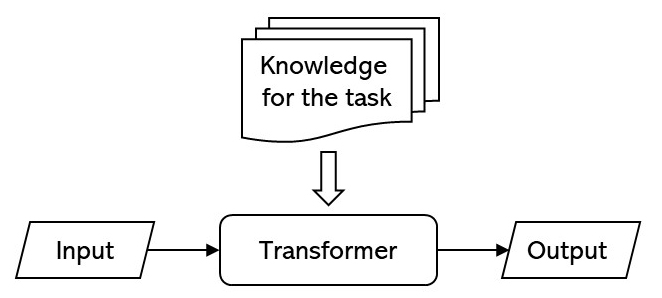}
	%\fbox{\rule[-.5cm]{4cm}{4cm} \rule[-.5cm]{4cm}{0cm}}
	\caption{Experiment Framework}
	\label{fig:fig3}
\end{figure}

\begin{table}[ht]
	\caption{Experiment Settings}
	\centering
	\begin{tabular}{p{2.5cm}p{5.8cm}p{4cm}p{2.5cm}}
		\toprule
		%\multicolumn{2}{c}{Part}                   \\
		%\cmidrule(r){1-2}
		Experiment     & Knowledge for the task     & Input    &Transformer\\
		\midrule
		Problem-driven reasoning   & RedDot award-winning design   & Problem statement, concept category     &Fine-tuned GPT-2\\
		\midrule
		Analogy-driven reasoning     & Examples of design-by-analogy concepts from RedDot award-winning design   & Target and source domains       &GPT-3 few-shot learning\\

		\bottomrule
	\end{tabular}
	\label{tab:table1}
\end{table}

\section{Experiments}
\label{sec:headings}
\subsection{Problem-Driven Reasoning}
Given GPT-2's capability to generate text based on understanding the context via training and fine-tuning, we experiment its application to generating text of solution ideas for a given problem. Problem-solving in design could be supported by different methods such as analogy and first principle. In this experiment we do not constrain GPT-2's problem-solving approach. The dataset for model fine-tuning is collected from RedDot's official website (https://www.red-dot.org/), including 14,502 product designs from 2011 to 2020 and 1,486 design concepts from 2016 to 2020. Data preparation includes picking out the text description of each design and adding its category name before the description. An example description of a problem-driven design in the RedDot dataset is shown below:
\begin{quote}
\textit{“One of the biggest and most common concerns of using public toilets is avoiding dermatosis and bacterial infection that comes from sharing a toilet with others. Clean Seat has a toilet lid that automatically opens when the first sensor (located at the front of the toilet lid) detects a user approaching. A second sensor then detects the person leaving after using the toilet, prompting the toilet lid to close and lock itself. When the lid is locked, the system kickstarts the self-cleaning function of the toilet."}
\end{quote}
As shown in the example, the problem is stated in the first sentence, followed by the solution idea description. This structure is common in the descriptions of problem-driven design and ideal for GPT to generate solution idea text as the output in response to a problem text as the input. However, not all award-winning designs are problem-driven and often the description does not begin with a problem description. Our hypothesis is that the model can learn from those problem-driven design descriptions (as the example above) in the total dataset to execute problem-solving tasks, while also learning from design descriptions of other structures during training. A pre-trained GPT-2 of 774M parameters is fine-tuned for 4,000 steps. Figure 4 reports average loss over the fine-tuning steps.
\begin{figure}[ht]
	\centering
	\includegraphics[width = 10cm]{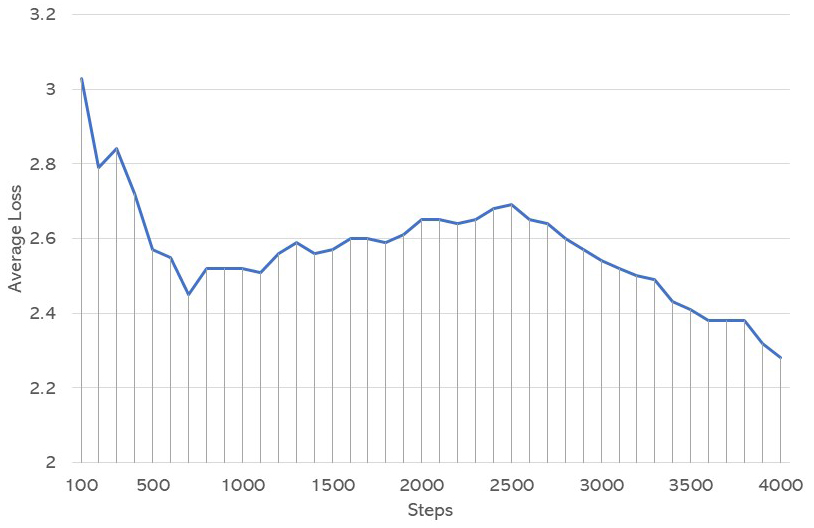}
	\caption{Average loss with training steps when fine-tuning with RedDot data}
	\label{fig:fig4}
\end{figure}
In the generation phrase, the problem statement and the design domain category are both set as input. The problem statement can be a single sentence written by the designer, and the category input can be selected from the categories listed on RedDot official website. Table 2 reports two solutions generated for each of the two design problems in two experiments (parameters: temperature = 0.85, top-k = 40, and top-p = 1). The first problem statement is from in the example above and the other is from an IF award-winning design (https://ifworlddesignguide.com/entry/324944-ema), in order to test the model's capability to generate solution ideas to known versus unknown problems to the model.
\begin{table}[ht]
	\caption{Problem-driven reasoning experiment: inputs and generated concepts}
	\centering
	\begin{tabular}{p{1.5cm}p{4cm}p{9.5cm}}
		\toprule
		\multicolumn{2}{c}{Input}                   \\
		\cmidrule(r){1-2}
		Category     & Problem Statement    & Generated Solutions\\
		\midrule
		Personal Hygiene   & \paragraph{} One of the biggest and most common concerns of using public toilets is avoiding dermatosis and bacterial infection that comes from sharing a toilet with others.   & \paragraph{Solution 1} Clean Seat provides a new and simple solution to this problem. A toilet seat equipped with an auto-level function is used to disinfect the seat after each use. The clean seat deodorises the entire seat, thus dramatically reducing the bacterial load on the back of the toilet bowl. Moreover, the seat has a crotch-level rubber band that provides comfort and personal comfort when using the toilet. This product is designed in a way that it provides a pleasant and personal experience using public toilets. The user can become part of the solution by donating their old clothes to the public toilet maintenance.\par \paragraph{Solution 2} To deal with this, the Trit objective was to design a toilet that is free of such infections. This was achieved through a very low-pollution level using of organic filter and antimicrobial ceramic material. The material is not only sustainable but is also highly hygienic. \\
		\midrule
		Life Science   & \paragraph{} Current electrocardiograph testing involves a large number of wires, making the process time-consuming for doctors and stressful for children.   & \paragraph{Solution 1} The suction coil is attached to the skin and the image data are transmitted to a tablet via Bluetooth. The spiral design of the coil makes it easy to position the device quickly on the skin during testing. An infrared sensor in the base of the device sends infrared light to the face and indicates the test status via a light ring. This allows doctors to locate the hygienic needle quickly in the event of a failed test result.\par \paragraph{Solution 2} The way out-of-office is also easier without these wires. With a simple hand-type electrocardiogram sensor, a doctor achieves the same results with a lower voltage and reduced energy consumption. \\
		\bottomrule
	\end{tabular}
	\label{tab:table2}
\end{table}
The fine-tuned GPT-2 generated many more concepts than the four demonstrated in Table 3, allowing us to evaluate its performance. For the known problem with a known solution, GPT-2 can generate many new solution ideas based on the massive knowledge it has acquired from the training. For the unknown problem, it is also capable of comprehending the problem and generating solution ideas. However, as the text lengthens, it is likely to start talking about other features of the design that are unrelated to the given problem. This is not surprising as the design descriptions in the fine-tuning dataset often provide comprehensive elaboration of multiple aspects and the model learnt this well.

\subsection{Analogy-Driven Reasoning}
Design-by-analogy is the projection of existing reference in a source domain to address a comparable challenge in the target domain \citep{gentner1983structure,luo2021guiding}. It is usually considered as a problem-solving approach. However, when a problem is not specified, analogy reasoning can also lead to open-ended design concept generation. This experiment is to test the analogy-driven reasoning of the model for concept generation, particularly for the context when a designer aims to draw analogy from a given source domain to generating design concepts in a given target domain but has not established clear analogy mapping across domains to generate specific new concepts.\par
As there are insufficient design-by-analogy examples to fine-tune a GPT-2, this experiment employs five analogy-driven reasoning examples selected from RedDot dataset as prompts for GPT-3's few-shot learning. Table 3 shows the source and target domains in each of the five examples for learning. Before each example is inputted, a structured sentence specifying the source and target domains (e.g., “Applying accordion to computer mouse”) is inserted so that the GPT-3 may learn to develop ideas based on the input domains. When generating new concepts, we simply need to update the tokens that specify the source and target domains in the input sentence.
\begin{table}[ht]
	\caption{Analogy-driven reasoning examples used as prompts for GPT-3}
	\centering
	\begin{tabular}{p{2.5cm}p{2.5cm}p{10cm}}
		\toprule
		%\multicolumn{2}{c}{Part}                   \\
		%\cmidrule(r){1-2}
		Source Domain     & Target Domain     & Link of the example\\
		\midrule
		Accordion   & Computer Mouse   & \url{https://www.red-dot.org/project/ambi-48504} \\
		\midrule
		Cells     & Building   & \url{https://www.red-dot.org/project/build-fender-27044}\\
		\midrule
		Standing desk   & Automobile   & \url{https://www.red-dot.org/project/sole-26525} \\
		\midrule
		Folding chair     & Wheelchair   & \url{https://www.red-dot.org/project/fold-light-wheelchair-26521}\\
		\midrule
		Circuit board     & Desk   & \url{https://www.red-dot.org/project/cabletread-46563}\\

		\bottomrule
	\end{tabular}
	\label{tab:table3}
\end{table}

Table 4 reports two drone design concepts generated by GPT-3, by drawing analogy from the given source domains of lantern and origami in two experiments (parameters: temperature = 0.85, top-p = 1, presence\_penalty=0.5, frequency\_penalty=0.5). The model successfully understands both source domains and applies the lighting feature of the lantern and the folding mechanism of origami to drone design. Without being trained with a large amount of analogical design cases, the generated concepts do adhere to analogical reasoning and learn to build a clear analogical mapping between the source and target domains, even when the two given source domains are culture-related and distant from the technical domain of drone. This is because the GPT-3 is pre-trained on a massive text dataset that includes nearly all possible kinds of scenarios and knowledges and a few examples as prompts are enough for the specific 'memory' of the model to be evoked.\par
Furthermore, consistent to the pervious experiment, we also tested for the model's performance with the input of the same prompt sentences as in the given examples. It turns out GPT-3 is more likely to repeat the ideas given in the examples given low frequency penalty and presence penalty parameters, but it begins to generate novel texts as the parameters are set to higher values. This reveals another difference between the mechanisms of GPT-2 and GPT-3 models when being applied to design concept generation tasks, which will need further investigation.

\begin{table}[ht]
	\caption{Analogy-driven reasoning experiment: inputs and generated concepts}
	\centering
	\begin{tabular}{p{1.3cm}p{1.3cm}p{12cm}}
		\toprule
		\multicolumn{2}{c}{Input}                   \\
		\cmidrule(r){1-2}
		Source Domain     & Target Domain   & Generated Concepts\\
		\midrule
		Lantern   & Drone    & \paragraph{Concept 1} Focused on the design of portable drone landing pads, Skylight is conceived as a portable lantern that allows operators to track flying drones for charging or repair. Developed into different designs that can fit with any situation, Skylight has two main forms: docking stations that are focused on portability and public space; and lamp-like structures for open areas. \par \paragraph{Concept 2} D-LIGHT is a decorative drone that is fitted with a lantern inspired by the Chinese ‘paper lantern’. The drone’s movements control the position and movement of the lantern, which is mounted at its centre. The functional design of D-LIGHT focuses on responsive mechanisms for flying and controlling the lantern. This also enables it to be operated wherever necessary.\\
		\midrule
		Origami   & Drone   & \paragraph{Concept 1} Aerobowl is inspired by the art of origami. The drone folds like an aerial paper aircraft and is able to fold itself up after landing. It can be launched without any human intervention, for example into lakes or rivers to help with rescue operations; it then folds its propellers into lightweight aircraft mode before floating along on the surface of the water like a boat. \par \paragraph{Concept 2} For centuries, paper folding has evolved into various different forms of art. Inspired by this, PaperDrone is an origami-inspired drone that is designed to have its chassis structure be deformable. The main goal of the design is to minimize the folding part while maintaining its strength and rigidity, mainly focusing on the legs for this feature. There are three feathered blades attached to each leg that can rotate in both directions at certain angles. The foldable body allows it to be very compact when folded, thus making easy storage possible. \\
		\bottomrule
	\end{tabular}
	\label{tab:table4}
\end{table}

\section{Discussion and Future Works}
\label{sec:headings}
This article has examined the uses of generative pre-trained transformers to generate design concepts. It is demonstrated that by customizing the training data or examples, GPT can perform conceptual design tasks with a reasonable level of competence. This work opens the path to verbal concept generation using NLG, which may possibly be integrated with other approaches for a broader variety of applications and a higher level of automation. For example, the analogy-driven reasoning experiment necessitates source domain data for concept generation and comparison. This may be supplemented with knowledge graphs that provide domain knowledges with a given knowledge distance. Furthermore, given suitable datasets, the generation of design concepts based on varied design heuristics will be possible, such as TRIZ \citep{ilevbare2013review} or the 77 design heuristics \citep{yilmaz2016evidence}.\par
However, it should be highlighted that the computer-generated concepts we showcased above were selected from a pool of low-quality results that may not be viable or context relevant. Thus, efficient concept evaluation algorithm is required to filter the automatically generated design concepts. Such an evaluator should be able to gauge both concept quality and novelty. Furthermore, human assessment should be employed to validate the performances of both the generator and evaluator.
\\
\\
\\
\\
\\

\bibliographystyle{unsrtnat}
\bibliography{references}  %%% Uncomment this line and comment out the ``thebibliography'' section below to use the external .bib file (using bibtex) .





% \end{thebibliography}

\end{document}